\documentclass[sigconf]{acmart}
\usepackage[utf8]{inputenc}

\usepackage[noend]{algpseudocode}
\usepackage{algorithm}
\usepackage{amssymb}
\usepackage{array}
\usepackage{changes}
\usepackage{comment}
\usepackage{esvect}
\usepackage{float}
\usepackage{graphicx}
\usepackage{mathtools}
\usepackage{multirow}
\usepackage{nicefrac}
\usepackage{url}
\usepackage{wrapfig}

\floatname{algorithm}{Procedure}

\usepackage[space]{grffile}

\definechangesauthor[color=red]{is}
\definechangesauthor[color=blue]{js}

\usepackage[colorinlistoftodos,prependcaption,textsize=tiny]{todonotes}

\newcommand{\expected}{\mathop{\mathbb{E}}}
\newcommand{\observed}{\mathbb{S}}
\newcommand{\estimated}{\widetilde{\mathbb{S}}}
\newcommand{\sample}{\mathord{\sim}}

\AtBeginDocument{%
  \providecommand\BibTeX{{%
    \normalfont B\kern-0.5em{\scshape i\kern-0.25em b}\kern-0.8em\TeX}}}

\setcopyright{acmcopyright}
\copyrightyear{2020}
\acmYear{2020}
\acmDOI{10.1145/1122445.1122456}

\acmConference[MLG'20]{MLG 2020: 16th International Workshop on Mining and Learning with Graphs}{August 24, 2020}{San Diego, CA}
\acmPrice{15.00}
\acmISBN{978-1-4503-XXXX-X/18/06}

\begin{document}

\title{First- and High-Order Bipartite Embeddings}

\author{Justin Sybrandt}
\thanks{Justin Sybrandt is now at Google Brain. Contact via \href{mailto:jsybrandt@google.com}{jsybrandt@google.com}.}
\email{jsybran@clemson.edu}
\orcid{0000-0001-5073-0122}
\affiliation{%
  \institution{Clemson University}
  \streetaddress{821 McMillan Rd.}
  \city{Clemson}
  \state{SC}
  \postcode{29631}
}

\author{Ilya Safro}
\email{isafro@clemson.edu}
\orcid{0000-0001-6284-7408}
\affiliation{%
  \institution{Clemson University}
  \streetaddress{821 McMillan Rd.}
  \city{Clemson}
  \state{SC}
  \postcode{29631}
}

\begin{abstract}
Typical graph embeddings may not capture type-specific bipartite graph 
features that arise in such areas as recommender systems, data visualization, 
and drug discovery. Machine learning methods utilized in these applications 
would be better served with specialized embedding techniques.
We propose two embeddings for bipartite graphs that decompose edges into sets
of indirect relationships between node neighborhoods. When sampling higher-order
relationships, we reinforce similarities through algebraic distance on graphs.
We also introduce ensemble embeddings to combine both into a ``best of both 
worlds'' embedding. The proposed methods are evaluated on link prediction and 
recommendation tasks and compared with other state-of-the-art embeddings. Our embeddings are found to perform better on recommendation tasks and equally competitive in link prediction. 
Although all considered embeddings are beneficial in particular applications, we demonstrate that none of those considered is clearly superior (in contrast to what is claimed in many papers). Therefore, we discuss the trade offs among them, noting that the methods proposed here are robust for applications relying on same-typed comparisons.\\
{\bf Reproducibility}: Our code, data sets, and results are all publicly
available online at: \url{https://sybrandt.com/2020/fobe_hobe/}.
\end{abstract}



\maketitle

\section{Introduction} \label{sec:introduction}

Graph embedding methods place nodes into a continuous vector space in order to
capture structural properties that enable machine learning
tasks~\cite{goyal2018graph}. While many have made significant progress embedding
general graphs~\cite{perozzi2014deepwalk, tang2015line,
grover2016node2vec, tsitsulin2018verse}, we find that bipartite graphs have received less
study~\cite{gao2018bine}, and that the field is far from settled on this interesting
case.
There exist a variety of special algorithmic cases for bipartite graphs, which are utilized in applications such as user-product or user-group recommender systems~\cite{zhang2017rerevisiting}, hypergraph based load balancing and mapping \cite{naumann2012combinatorial}, gene-disease relationships~\cite{barabasi2011network}, and drug-to-drug  targets~\cite{yildirim2007drug}.

We define a simple, undirected, and unweighted bipartite graph to be $G=(V,E)$ where $V=\{v_1,v_2,\dots,v_{n+m}\}$ is composed of the \emph{disjoint} subsets $A=\{\alpha_1,\dots,\alpha_n\}$ and $B=\{\beta_1,\dots,\beta_n\}$ ($V=A\cup B$). Here, $A$ and $B$ represent the two halves of the network, and are sometimes called ``types.''
We use $v_i$ to indicate any node in $V$, $\alpha_i$ for nodes in $A$, and $\beta_i$ for those in $B$.
In a bipartite graph, edges only occur across types, and $E \subseteq\{A \times B\}$ indicates those  connections within $G$.
A single edge is notated as $\alpha_i\beta_j \in E$, and because our graph is undirected, $\alpha_i\beta_j = \beta_j\alpha_i$.
The neighborhood of a node is indicated by the function $\Gamma(\cdot)$.
If $\alpha_i \in A$ then ${\Gamma(\alpha_i)=\{\beta_j | \alpha_i\beta_j \in E\}}$, and vice-versa for nodes in $B$.
In order to sample an element from a set, such as selecting a random $\alpha_i$ from $A$ with uniform probability, we notate $\alpha_i \sample A$.
The problem of graph embedding is to determine a representation of the nodes in $G$ in a vector space of $r$ dimensions such that $r << |V|$ and that a select node-similarity measure defined on $V$ is encoded by these vectors~\cite{tsitsulin2018verse}.
We notate this embedding as the function $\epsilon(\cdot):V\rightarrow \mathbb{R}^r$, that maps each node to an embedding.

We propose two methods for embedding bipartite graphs. These
methods fit embeddings by optimizing nodes of each type separately, which we
find can lead to higher quality type-specific latent features. Our first method,
First-Order Bipartite Embedding (FOBE), samples for the existence of
direct, and first-order similarities within the bipartite structure. This
approach maintains the separation of types by reformulating edges in $E$
into indirect same-typed observations.
For instance, the connection $\alpha_i\beta_j \in E$ decomposes into a set of observed pairs $(\alpha_i, \alpha_k \sample \Gamma(\beta_j))$ and 
$(\beta_j, \beta_k \sample \Gamma(\alpha_i))$.

Our second method, High-Order Bipartite Embedding (HOBE), samples direct,
first-, and second-order relationships, and weighs samples
using algebraic distance on bipartite graphs~\cite{chen2011algebraic}.
Again, we represent sampled relationships between nodes of different types by
decomposing them into collections of same-typed relationships.
While this sampling approach is similar to FOBE, algebraic distance allows us to
improve embedding quality by accounting for broader graph-wide
trends. Algebraic distance on bipartite graphs has the effect of capturing
strong local similarities between nodes, and reduces the effect of less
meaningful relationships.
This behavior is beneficial in
many applications, such as shopping, where two users are likely more
similar if they both purchase a niche hobby product, and may not be similar even
if they both purchase a generic cleaning product.

Because FOBE and HOBE each make different prior assumptions about the relevance
of bipartite relationships, we propose a method for combining bipartite
embeddings to get ``best of both worlds'' performance. This ensemble approach
learns a joint representation from multiple pre-trained embeddings.  The
``direct'' combination method fits a non-linear transformation of the original
embeddings into a fixed-size hidden layer in accordance to sampled similarities.
The ``auto-regularized'' combination extends the direct method by introducing a
denoising-autoencoder layer in order to regulate the learned joint
embedding~\cite{vincent2008extracting}. The architecture of both approaches
maintains a separation between nodes of different types, which allows for
type-specific embeddings, without the constraint of a shared global structure. Evaluation of all proposed  embeddings is performed on link prediction reinforced with holdout experiments and recommender system tasks.

\paragraph{Our contribution in summary:} (1) We introduce First- and High-Order Bipartite Embeddings that learn
representations of bipartite structure that retaining
type-specific semantic information. (2) We present the direct and the auto-regularized methods to leverage
multiple pre-trained graph embeddings to produce a ``best of both
words'' embedding. (3) We discuss the strengths and weaknesses of our proposed methods as
they compare to a range of graph embedding techniques. We identify certain graph
properties that suit different graph types, and report that none of the proposed
embeddings is clearly superior. However, we find that applications wanting to
make many same-typed comparisons are often best suited by a type-sensitive
embedding. 

\subsection{Related Work} \label{sec:background}

Low-rank embeddings project 
high-order data into a compressed real-valued space, often for the purpose of facilitating machine learning models.
Inspired by the Skip-Gram approach\cite{mikolov2013efficient}, Perozzi et al. demonstrate that for a
similar method can capture latent structural features of traditional
graphs~\cite{perozzi2014deepwalk}. 
An alternative approach, LINE by Tang et al., models first- and second-order
node relationships explicitly~\cite{tang2015line}. 
Node2Vec blends the intuitions behind both LINE and Deepwalk by combining
homophilic and structural similarities through a biased random
walk~\cite{grover2016node2vec}. 
Our proposed
methods are certainly influenced by LINE's approach, but differ in a few key
areas. Firstly, we split our model in order to only make same-typed comparisons.
Furthermore, we introduce terms that compare nodes with relevant negihborhoods, and can weigh different samples with algebraic distance~\cite{chen2011algebraic}.

While the three previously listed embedding approaches are designed for
traditional graphs, Metapath2Vec++ by Dong et al.  presents a heterogeneous
approach using extended type-sensitive skip-gram model~\cite{dong2017metapath2vec}.
Our method differs from Dong et al.'s in a number of
ways.  Again, we do not apply random walks or the skip-gram model.  Furthermore,
the Metapath2Vec++ model implicitly asserts that output type-specific embeddings
be a linear combination of the same hidden layer. In contrast, we create
entirely separate embedding spaces for the nodes of  different types.
BiNE by Gao et al. focuses directly on the
bipartite case~\cite{gao2018bine}. This approach uses the biased random-walks
described in Node2Vec, and samples these walks in proportion to each node's HITS
centrality~\cite{kleinberg1999authoritative}.
While our methods differ, again, in the
use of skip-gram, BiNE also fundamentally differs from our proposed approaches
by enforcing global structure through cross-type similarities. 
Tsitsulin et al. present VERSE, a versatile graph embedding method that allows 
multiple different node-similarity measures to be captured by the same 
overarching embedding technique~\cite{tsitsulin2018verse}. This method requires
that the user specify a node-similarity measure that will be encoded in the
dot product of resulting embeddings. 
A key difference between the methods
presented here, and the methods presented in VERSE, come from differences in objective values when training embeddings.
VERSE uses a range of methods to sample node-pairs, from direct
sampling to Noise Contrastive Estimation~\cite{gutmann2010noise}, and 
updates embeddings according to their observed similarity or dissimilarity (in
the case of negative samples). However, the optimization method proposed
here enforces only same-typed comparisons.
\section{Methods and Technical Solutions}%
\label{sec:methods}

We present two sibling strategies for learning bipartite
embeddings.  First-Order Bipartite Embedding (FOBE) samples direct links from
$E$ and first-order relationships between nodes sharing common neighbors. We
then fit embeddings to minimize the KL-Divergence between our observations and
our embedding-based estimations. The second method, High-Order Bipartite
Embedding (HOBE), begins by computing algebraic similarity estimates for each edge
~\cite{chen2011algebraic,shaydulin2019relaxation}. Using these heuristic weights, HOBE
samples direct, first- and second-order relationships, to
which we fit embeddings using mean-squared error.

At a high level, both embedding methods begin by observing structural
relationships within a graph $G$ and then fitting an embedding $\epsilon$ in order to encode structural features via dot product of embeddings.
We combine three types of observations for a single graph
These observations are represented through the
functions $\observed_A(\cdot,\cdot)$,
$\observed_B(\cdot,\cdot)$, 
and $\observed_V(\cdot,\cdot)$.
Each function maps two nodes to an observed similarity: $V \times V \rightarrow \mathbb{R}$.
The result of $\observed_A$ is nonzero only if both arguments are in $A$, $\observed_B$ is similarly nonzero only if both arguments are in $B$. In this manner, these functions capture type-specific similarities. The $\observed_V$ function, in contrast, captures cross-typed observations, and is nonzero if its arguments are of different types.
We define a reciprocal set of functions to model these similarities: $\estimated_A(\cdot,\cdot)$, $\estimated_B(\cdot,\cdot)$,
and $\estimated_V(\cdot,\cdot)$. These functions are defined in terms of $\epsilon(\cdot)$, and each method must select some embedding such that the difference between each corresponding set of $\observed,\estimated$ pairs.

Because we estimate similarities within 
type-specific subsets of $\epsilon$ separately, we can better preserve typed latent features.  This is
important for many applications. Consider an embedding of the bipartite graph of
viewers and movies, often used for applications such as video recommendations.
Within ``movie space'' one would expect to uncover latent features such as
genre, budget, or the presence of high-profile actors. These features are
undefined within ``viewer space,'' wherein one would expect to observe latent
features corresponding to demographics and viewing preferences. Clearly these
two spaces are correlated in a number of ways, such as the alignment between
viewer tastes and movie genres. However, we find methods that enforce direct
comparisons between viewer and movie embeddings can result in an erosion of
type-specific features, which can lead to lower downstream performance.  In
contrast, the methods proposed here do not encode cross-type relationships as 
a linear transformation of embeddings, and instead captures cross-typed 
relationships through the aggregate behavior of node neighborhoods within
same-typed subspaces.

\subsection{First-Order Bipartite Embedding}

The goal of FOBE is to model direct and first-order relationships
from the original structure.
This very simple method only detects the existence of a relationship between two nodes, and therefore does not distinguish between two nodes that share only one neighbor from two nodes that share many. However, we find that this simplicity enables scalability at little cost to quality.
Here, a direct relationship is any edge from the
original bipartite graph, while a first-order relationship is defined as
$\{(\alpha_i,\alpha_j)\mid\Gamma(\alpha_i) \cap \Gamma(\alpha_j) \neq \emptyset\}$.
Note that nodes in a
first-order relationship share the same type. We define observations
corresponding with each relationship. Direct observations simply detect
the presence of an edge, while first-order relationships similarly detect a common neighbor. Formally:
\begin{equation}
  \observed_A(\alpha_i, \alpha_j) =
  \begin{cases}
    1 & \alpha_i,\alpha_j\in A \And
        \Gamma(\alpha_i) \cap \Gamma(\alpha_j) \neq \emptyset \\
    0 & \text{otherwise}
  \end{cases}
\end{equation}
\begin{equation}
  \observed_B(\beta_i, \beta_j) =
  \begin{cases}
    1 & \beta_i,\beta_j\in B \And
        \Gamma(\beta_i) \cap \Gamma(\beta_j) \neq \emptyset \\
    0 & \text{otherwise}
  \end{cases}
\end{equation}
\begin{equation}
  \observed_V(\alpha_i, \beta_j) =
  \begin{cases}
    1 & \alpha_i\beta_j\in E \\
    0 & \text{otherwise}
  \end{cases}
\end{equation}

By sampling $\gamma$ neighbors, we allow our later embedding model to
approximate the effects of $\Gamma$, similar to the $k$-ary set sampling
in~\cite{murphy2018janossy}. Note also that each sample contains one nonzero
$\observed$ value. By fitting all three observations
simultaneously, we implicitly generate two negative samples for each positive
sample.  Furthermore, we generate a fixed number of samples for each node's
direct and first-order relationships.

Given these observations $\observed_A$, $\observed_B$, and $\observed_V$, we fit the $\epsilon$ embedding
according to corresponding estimation functions $\estimated_A$, $\estimated_B$, $\estimated_V$.  To estimate a
first-order relationship ($\estimated_A$ and $\estimated_B$) we calculate the sigmoid of the
dot product of embeddings (\ref{eq:bhbe:first_order}), namely,
\begin{equation}
  \sigma(x) = (1+e^{-x})^{-1}.
\end{equation}
\begin{equation}
  \estimated_A(\alpha_i, \alpha_j) =
    \sigma\left(\epsilon(\alpha_i)^\intercal\epsilon(\alpha_j)\right)
  \label{eq:bhbe:first_order}
\end{equation}
\begin{equation}
  \estimated_B(\beta_i, \beta_j) =
    \sigma\left(\epsilon(\beta_i)^\intercal\epsilon(\beta_j)\right)
  \label{eq:bhbe:first_order:beta}
\end{equation}

Building from this, we train embeddings based on direct relationships by
composing relevant first-order relationships. Specifically, if $\alpha_i\beta_j\in E$
then we would expect $\alpha_i$ to be similar to $\alpha_k\in\Gamma(\beta_j)$ and vice-versa.
Intuitively, a viewer has a higher chance of watching a movie if they are
similar to others that have.  We formulate our direct relationship estimate to
be the product of each node's average first-order estimate to the other's
neighborhood. Formally:
\begin{equation}
  \estimated_{V}(\alpha_i, \beta_j) =
  \expected_{\alpha_k \in \Gamma(\beta_j)}\left[\estimated_A(\alpha_i, \alpha_k)\right]
  \expected_{\beta_k \in \Gamma(\alpha_i)}\left[\estimated_B(\beta_j, \beta_k)\right]
  \label{eq:bhbe:direct}
\end{equation}

In order to train our embedding function $\epsilon$ for the FOBE method, we
minimize the KL-Divergence~\cite{kullback1951information} between our observed
similarities $\observed$ and our estimated similarities $\estimated$. We minimize for
each simultaneously, for both direct and first-order similarities, using the
Adagrad optimizer~\cite{duchi2011adaptive}, namely, we solve:
\begin{equation}
\min_{\epsilon}
\sum_{v_i, v_j \in V \times V}\left[
\begin{aligned}
  \estimated_A(v_i, v_j)
  \log\left(
    \frac{\observed_A(v_i, v_j)}{\estimated_A(v_i, v_j)}
  \right) \\
  +
  \estimated_B(v_i, v_j)
  \log\left(
    \frac{\observed_B(v_i, v_j)}{\estimated_B(v_i, v_j)}
  \right) \\
  +
  \estimated_V(v_i, v_j)
  \log\left(
    \frac{\observed_V(v_i, v_j)}{\estimated_V(v_i, v_j)}
  \right)
\end{aligned}
\right]
\end{equation}



\subsection{High-Order Bipartite Embedding}%
\label{sec:methods:Algebraic_hypergraph2vec}

The goal of HOBE is to capture distant relationships between nodes that are related, but may not share an edge or a neighborhood.
In order to differentiate the meaningful distant connections from those that are spurious, we turn to algebraic distance on graphs~\cite{shaydulin2019relaxation}.
This method is fast to calculate and provides a strong signal for \emph{local similarity}.
For example, algebraic distance can tell us which neighbor of a high-degree node is the most similar to the root.
As a result, we can utilize this signal to estimate which multi-hop connections are the most important to preserve in our embedding.

Algebraic distance is a measure of dependence between variables popularized in algebraic multigrid (AMG) \cite{ron2011relaxation,BrandtBKL11,livne2012lean}. Later, it has been shown to be a reliable and fast way to capture implicit similarities between nodes in graphs \cite{safro:spars,leyffer2013fast} and hypergraphs that are represented as bipartite graphs \cite{shaydulin2019relaxation} (which is leveraged in this paper) taking into account distant neighborhoods. Technically, it is a process of relaxing randomly initialized test vectors using stationary iterative relaxation applied on graph Laplacian homogeneous system of equations, where in the end the algebraic distance between system's variables $x_i$ and $x_j$ (that correspond to linear system's rows $i$ and $j$) is defined as an maximum absolute value between the $i$th and $j$th components of the test vectors (or, depending on application, as sum or sum of squares of them).

In our context, a variable is a node, and we apply $K$ iterations of Jacobi over-relaxation (JOR) on the bipartite graph Laplacian as in \cite{ron2011relaxation} ($K=20$ typically ensures good stabilization as we do not need full convergence, see  Theorem 4.2 \cite{chen2011algebraic}).
Initially, each node's coordinate is assigned a random value, but on each
iteration a node's coordinate is updated to move it closer its neighbors'
average.  Weights corresponding to each neighbor are inversely proportional
their degree in order to increase the ``pull'' of small communities.
Intuitively, this acknowledges that two viewers who both watch a niche new-wave
movie are more likely similar than two viewers who watched a popular
blockbuster.  We run JOR on $R$
independent trials (called test  vectors in AMG works, convergence proven
in~\cite{chen2011algebraic}).  Formally, for $r$th test vector $a_r$  the update step of JOR 
is performed as follows, where $a_r^{(t)}(v_i)$ represents node $v_i$'s algebraic
coordinate on iteration $t\in \{1,..,K\}$, and $\lambda$ is a damping factor (suggested
$\lambda=0.5$ in~\cite{shaydulin2019relaxation}).
\begin{equation}
  {\bf a}_r^{(t+1)}(v_i) = \lambda {\bf a}_r^{(t)}(v_i)
               + (1-\lambda) \frac{
                 \sum_{v_j\in\Gamma(v_i)} {\bf a}_r^{(t)}(v_j)|\Gamma(v_j)|^{-1}
               }{
                 \sum_{v_j\in\Gamma(v_i)} |\Gamma(v_j)|^{-1}
               }
\end{equation}

We use the $l^2$-norm in order to summarize the algebraic distance of two nodes
across $R$ trails with different random initializations. 
As a result, two nodes will be close in our distance calculation if they remain
nearby across many trials, which lessens the effect of too slow convergence in a
single trial. For our purposes we select $R=10$. Additionally, we define
``algebraic similarity'', $s(i,j)$, as a closeness across trials. We subtract the distance
between two embeddings from the maximum distance in our space, and rescale the
result to the unit interval. Because we know that the maximum distance between
any two coordinates in the same trial is $1$, we can compute this in constant
time:
\begin{equation}
  d(v_i, v_j) = \sqrt{\sum_{r=1}^R \left(
                      {\bf a}_r^{(K)}(v_i) - {\bf a}_r^{(K)}(v_j)
                      \right)^2}
\end{equation}
\begin{equation}
  s(v_i, v_j) = \frac{\sqrt{R} - d(v_i, v_j)}{\sqrt{R}}
\end{equation}

After calculating algebraic similarities for pairs of nodes of all edges, we begin to sample direct, first-order,
and second-order similarities from the bipartite structure. Here, a second-order
connection is one wherein $\alpha_i$ and $\beta_j$ share a neighbor that shares a
neighbor: $\alpha_i\in\Gamma(\Gamma(\Gamma(\beta_j)))$. Note that the set of second-order
relationships is a superset of the direct relationships. We can extend to these
higher-order connections with HOBE, as opposed to FOBE, because of the
information provided in algebraic distances.  Many graphs contain a small
number of high degree nodes, which creates a very dense second-order graph.
Algebraic distances are therefore needed to distinguish which of the sampled
second-order connections are meaningful, especially when the refinement is
normalized by $|\Gamma(v_i)|^{-1}$.

We formulate our first-order observations to be equal to the strongest shared
bridge between two nodes. This indicates that both nodes are closely related
to something that is mutually representative, such as two viewers that watch
new-wave cinema. 
Formally:
\begin{equation}
  \observed_A^{'}(\alpha_i, \alpha_j) =
  \begin{cases}
    \displaystyle\max_{
        \beta_k\in\Gamma(\alpha_i)\cap\Gamma(\alpha_j)
    }
    \min\left(
        s(\alpha_i, \beta_k),
        s(\alpha_j, \beta_k)
    \right)\\
    \quad\text{if}~\alpha_i,\alpha_j\in A \\
    0~\text{otherwise}
  \end{cases}
\end{equation}
\begin{equation}
  \observed_B^{'}(\beta_i, \beta_j) =
  \begin{cases}
    \displaystyle\max_{
        \alpha_k\in\Gamma(\beta_i)\cap\Gamma(\beta_j)
    }
    \min\left(
        s(\alpha_k, \beta_i),
        s(\alpha_k, \beta_j)
    \right) \\
    \quad\text{if}~\beta_i,\beta_j\in B \\
    0~\text{otherwise}
  \end{cases}
\end{equation}

When observing second-order relationships between nodes $\alpha_i$ and $\beta_j$ if different types, we again construct a measurement from shared
first-order relationships. Specifically, we are looking for the strongest
first-order connection between $i$ and $j$'s neighborhood, and vice-versa.
In the context of viewers and movies this represents the similarity between a
viewer and a movie watched by a friend. Formally:
\begin{equation}
  \observed_{V}^{'}(\alpha_i, \beta_j) = \max\left(
  \begin{aligned}
    \max_{\alpha_k \in \Gamma(\beta_j)}
        \observed_A^{'}(\alpha_i, \alpha_k),
    \max_{\beta_k \in \Gamma(\alpha_i)}
        \observed_B^{'}(\beta_j, \beta_k)
  \end{aligned}
    \right)
\end{equation}

We again collect a fixed number of samples for each relationship type: direct,
first- and second-order. 
We then train embeddings using cosine similarities,
however we select the ReLU activation function to replace sigmoid in
order to capture the weighted relationships. We optimize for all three
observations simultaneously, which again has the effect of creating negative
samples for non-observed phenomena. Our estimated similarities are defined as
follows:
\begin{equation}
  \estimated_A^{'}(\alpha_i, \alpha_j) =
    \max\left(0, \epsilon(\alpha_i)^\intercal\epsilon(\alpha_j)\right)
\end{equation}
\begin{equation}
  \estimated_B^{'}(\beta_i, \beta_j) =
    \max\left(0, \epsilon(\beta_i)^\intercal\epsilon(\beta_j)\right)
\end{equation}
\begin{equation}
  \estimated_{B}^{'}(\alpha_i, \beta_j) =
  \expected_{\alpha_k \in \Gamma(\beta_j)}\left[
    \estimated_A^{'}(\alpha_i, \alpha_k)
  \right]
  \expected_{\beta_k \in \Gamma(\alpha_i)}\left[
    \estimated_B^{'}(\beta_j, \beta_k)
  \right]
\end{equation}

We use the same model as FOBE to train
HOBE, but with our new estimation functions and a new objective.  We now
optimize for the mean-squared error between our observed and estimated samples,
as KL-Divergence is ill-defined for the weighted samples we collect.  Formally,
we minimize:
\begin{equation}
  \min_\epsilon \expected_{v_i, v_j \in V \times V}
  \left[
  \begin{aligned}
    (\observed_A^{'}(v_i, v_j) - \estimated_A^{'}(v_i, v_j))^2 \\
    + (\observed_B^{'}(v_i, v_j) - \estimated_B^{'}(v_i, v_j))^2 \\
    + (\observed_V^{'}(v_i, v_j) - \estimated_V^{'}(v_i, v_j))^2 \\
  \end{aligned}
  \right]
\end{equation}

\subsection{Combination Bipartite Embedding}%
\label{sec:methods:ensemble_graph_embedding}

In order to unify our proposed approaches, we
present a method to create a joint embedding from multiple pre-trained
bipartite embeddings. This combination method maintains our
initial assertion that nodes of different types ought to participate in
different global embedding structures.
We fit a non-linear projection of the input embeddings such that an intermediate
embedding can accurately uncover direct relationships. This raises a question as
to whether it is better to create an intermediate that succeeds in this training
task, or whether it is better to fully encode the input embeddings. To address
this concern we propose two flavors of our combination method: the ``direct''
approach maximizes performance on the training task, while the
``auto-regularized'' approach enforces a full encoding of input embeddings.

We begin by
taking the edge list of the original bipartite graph $E$ as our set of positive
samples. We then generate five negative samples for each node by selecting
random pairs $\alpha_i\beta_j \notin E$.
For each sample, we create an input vector by concatenating each of the $e'$ pre-trained embeddings.
\begin{equation}
  In(v_i) = \left[\epsilon_1(v_i)~~\epsilon_2(v_i)~~...~~\epsilon_{e'}(v_i)\right]
\end{equation}

After generating $In(\alpha_i)$ and $In(\beta_j)$, our models assert 50\% dropout in
these input vectors~\cite{srivastava2014dropout}. We do so in the
auto-regularized case so that we follow the pattern of denoising auto-encoders,
which have shown high performance in robust dimensionality
reductions~\cite{vincent2008extracting}. However, we also find that this dropout
increases performance in the direct combination model as well. This is because
in either case, we anticipate both redundant and noisy signals to be present
across the concatenated embeddings. This is especially necessary for larger
values of $k$ and $e'$, where the risk of overfitting increases.

We then project $In(\alpha_i)$ and $In(\beta_j)$ separately onto two hidden layers of
size $\nicefrac{d(In)+k^{'}}{2}$ where $d(\cdot)$ indicates the dimensionality of the input,
and $k^{'}$ represents the desired dimensionality of the combined embeddings. By
separating these hidden layers, we only allow signals from within embeddings of
the same node to affect its combination. We then project down to two combination
embeddings of size $k^{'}$, which act as input to both the joint link-prediction
model, as well as to the optional auto-encoder layers.

In the direct case, we simply minimize the mean-squared error between the
predicted links and the observed links. Formally, let
${\observed^{''}(\alpha_i,\beta_j)\rightarrow\{0,1\}}$ equal the sampled value, and let
${\estimated^{''}(\alpha_i,\beta_j)\rightarrow\mathbb{R}}$ be combination estimate. 
In the auto-regularized case we introduce a factor to enforce that the original
(pre-dropout) embeddings can be recovered from the combined embedding. We weight
these factors so they are half as important as performing the link
prediction training task. The neural architecture used to learn these combination embeddings is depicted in the supplemental information. If $\Theta$ is the set of free parameters of our neural network model, $N$ is the set of negative samples, and $Out(v_i)$ is the output of the auto-encoder corresponding to $In(v_i)$, then we optimize the following (direct followed by auto-regularized):
\begin{equation}
  \min_\Theta \expected_{\alpha_i, \beta_j \in (E+N)} \left(\observed^{''}(\alpha_i, \beta_j) - \estimated^{''}(\alpha_i, \beta_j)\right)^2
\end{equation}
\begin{equation}
  \min_\Theta
    \expected_{\alpha_i, \beta_j \in (E+N)} \left(
\begin{aligned}
      4 \left(\observed^{''}(\alpha_i, \beta_j) - \estimated^{''}(\alpha_i, \beta_j)\right)^2 \\
      + ||In(\alpha_i) - Out(\alpha_i)||_2 \\
      + ||In(\beta_j) - Out(\beta_j)||_2
\end{aligned}
    \right)
\end{equation}

\section{Algorithmic Analysis}
In order to efficiently compute FOBE and HOBE, we collect a fixed number of samples per node for each of the observation functions, $\observed$.
As later explored in Table~\ref{tab:sensitivity_study}, we find that the performance of our proposed methods does not significantly increase beyond a relatively small, fixed sampling rate $s_r$, where $s_r << |V|$.
Using this observation, we can efficiently minimize the FOBE and HOBE objective values by approximating the expensive $O(n^2)$ set of comparisons ($v_i,v_j\in V\times V$) with a linear number of samples (specifically $O(|V|s_r)$).
Furthermore, we can estimate the effect of each node's neighborhood in observations $\observed_V$ and $\observed_V'$ by following a similar approach.
Instead of considering each node's total $O(V)$-sized neighborhood, we can randomly sample $s_\gamma$ neighboring nodes with replacement.
These specifically samples nodes are recorded during the sampling procedure so that they may be referenced during training.
Algorithm~\ref{alg:fobe_hobe_sampling} describes the sampling algorithm formally.
\begin{algorithm}[t]
\begin{algorithmic}[1]
\Function{SameTypeSample}{$v_i, s_r, \observed$}
    \State $v_j \sample \Gamma(\Gamma(v_i))$
    \State Record $v_i$,$v_j$, and $\observed(v_i,v_j)$ 
\EndFunction

\Function{DiffTypeSample}{$v_i, s_r, s_\gamma, G, \observed$}
    \State $v_j \sample G(v_i)$
    \State Let $\gamma_\alpha$ and $\gamma_\beta$ be sets of size $s_\gamma$ sampled with replacement from the neighborhoods $\Gamma(v_i)$ and $\Gamma(v_j)$ according to the types of $v_i$ and $v_j$.
    \State Record $v_i, v_j, \gamma_\alpha, \gamma_\beta,$ and $\observed(v_i, v_j)$.
\EndFunction

\Function{FobeSampling}{$G, s_r, s_\gamma$}
    \ForAll{$v_i \in V$}
        \For{$s_r$ samples}
            \State\Call{SameTypeSample}{$v_i, s_r, \observed_A$}
            \State\Call{SameTypeSample}{$v_i, s_r, \observed_B$}
            \State\Call{DiffTypeSample}{$v_i, s_r, s_\gamma, \Gamma(\cdot), \observed_V$}
        \EndFor
    \EndFor
\EndFunction

\Function{HobeSampling}{$G, s_r, s_\gamma$}
    \ForAll{$v_i \in V$}
        \For{$s_r$ samples}
            \State\Call{SameTypeSample}{$v_i, s_r, \observed_A'$}
            \State\Call{SameTypeSample}{$v_i, s_r, \observed_B'$}
            \State\Call{DiffTypeSample}{$v_i, s_r, s_\gamma, \Gamma(\Gamma(\Gamma(\cdot))), \observed_V'$}
        \EndFor
    \EndFor
\EndFunction
\end{algorithmic}
\caption{FOBE/HOBE Sampling. Unobserved values per sample are recorded as either zero or empty.}
\label{alg:fobe_hobe_sampling}
\end{algorithm}
\section{Empirical Evaluation} \label{sec:experiments}

\noindent{\bf{Link Prediction}}
We evaluate the performance of our proposed embeddings across three
link prediction tasks and a range of training-test splits.
When removing edges, we visit each in random order and remove them with probability $h$ provided the removal does not disconnect the graph.
This additional check ensures all
nodes appear in all experimental embeddings. The result is the subgraph $G'=(V,
E', h)$. Deleted edges form the positive test-set examples, and we generate set of negative samples (edges not present in original graph) of equal size.
These samples are used to
train three sets of link-prediction models: the $A$-Personalized,
$B$-Personalized (where $A$ and $B$ are parts of $V$), and unified models.

The $A$-personalized model is a support vector machine trained on the neighborhood of a particular node.
A model personalized to $i \in A$ learns to
identify a region in $B$-space corresponding to its neighborhood in $G'$.
We use support vector machines with the radial basis
kernel ($C=1,\gamma=0.1$) because we find these models result in robust
performance given limited training data, and because the chosen kernel function
allows for non-spherical decision boundaries. We additionally generate five negative samples for each positive sample (a neighbor of $i$ in $G'$). In doing so we evaluate the ability to capture type-specific latent features, as each personalized model only considers one-type's embeddings.
While the personalized task may not be typical for production link-prediction systems, it is
an important measure of latent features found in each space.
In many bipartite applications, such as the
six we have selected for evaluation, $|A|$ and $|B|$ may be drastically
different. For instance, there are typically more viewers than movies, or more buyers than products.
Therefore it becomes important to understand the differences in quality between the latent spaces of each type, which we evaluate through  these personalized models.

The unified link-prediction model, in contrast, learns to associate $\alpha_i\beta_j\in E'$
with a combination of $\epsilon(\alpha_i)$ and $\epsilon(\beta_j)$.
This model attempts to quantify global trends across embedding spaces.
We use a hidden layer of size $k$ with the ReLU activation function, and a single
output with the sigmoid activation. We fit this model against mean-squared error
using the Adagrad optimizer \cite{duchi2011adaptive}.

{\bf Datasets.}
We evaluate each embedding across six datasets.
The Amazon, YouTube, DBLP, Friendster, and
Livejournal graphs are all taken from the Stanford Large Network Dataset
Collection (SNAP)~\cite{leskovec2015snap}. We select the distribution of each
under the listing ``Networks with Ground-Truth Communities.''
Furthermore, we collect
the MadGrades graph, from an online source provided by the University of
Wisconsin at Madison~\cite{madgrades2018}.  This graph consists of teachers
and course codes, wherein an edge signifies that teacher $\alpha_i$ has taught course
code $\beta_j$. We clean this dataset by iteratively deleting any
instructor or course with degree $1$ until none remain.

{\bf Experimental Parameters.}
We evaluate the performance of our proposed methods: FOBE and HOBE, as well as
our two combination approaches: Direct and Auto-Regularized Combination
Bipartite Embedding. We compare against all methods described in Section~\ref{sec:background}.
Note, we limit our comparison to other embedding-based techniques as prior work~\cite{grover2016node2vec} establishes they considerably outperform alternative heuristic methods.
We evaluate each across the six above graphs and
nine training-test splits $h=0.1, 0.2, ..., 0.9$. 
For all embeddings we select dimensionality $k=100$.
For Deepwalk, we select a walk length of $10$, a window size of $5$, and $100$
walks per node. For LINE we apply the model that combines both first- and
second-order relationships, selecting 10,000 samples total and 5 negative
samples per node. For Node2Vec we select 10 walks per node, walk length
of $7$ and a window size of $3$. Furthermore, we select default parameters
for BiNE and Metapath2Vec++. For the latter, we supply the metapath of
alternating $A-B-A$ nodes, the only metapath in our bipartite case.
For FOBE and HOBE we generate $200$ samples per node, and when sampling
neighborhoods we select $5$ nodes with replacement upon each observation. After
training both methods, we fit the Direct and Auto-Regularized Combination
methods, each trained using \emph{only} the results of FOBE and HOBE.

\noindent{\bf{Recommendation}}:
We follow the procedure originally described by Gao et al. and evaluate our proposed embeddings through the task of recommendation~\cite{gao2018bine}.  Recommendation systems propose products to users in order to maximize the overall interaction rate. These systems fit the bipartite graph model because they are defined on the set of user-product interactions.
While many such systems could be reformulated as operations on bipartite networks, methods such as matrix factorization and user-user nearest neighbors do not capture granular local features to the same extent as modern graph embeddings~\cite{gao2018bine,bobadilla2013recommender}.
In contrast, bipartite graph embedding provides a framework to often learn richer latent representations for both users and products. These representations can then be used directly through simple similarity measures, or added to existing solution archetypes, such as k-nearest neighbors, which often provides significant quality benefits.

While there are many similarities between recommendation and link prediction, the key difference is the introduction of weighted connections. As a result, recommendation systems are evaluated based on their ability to rank products in accordance to held-out user supplied rankings. This is quantified through a number of metrics defined on the top $k$ system-supplied recommendation for each user.
When using embeddings to make a comparison, Gao et al. rank products by their embedding's dot product with a given user. However, our proposed methods relax the constraint that products and users be directly comparable. As a result, when ranking products for a particular user for our proposed embeddings we must first define a product-space representation. For each user we collect the set of known product ratings, and calculate a product centroid weighted by those ratings. 

{\bf Experimental Procedure.} We present a comparison between our proposed methods and all previously discussed embeddings across the DBLP%
\footnote{\url{https://github.com/clhchtcjj/BiNE/tree/master/data/dblp}}
and LastFM%
\footnote{\url{https://grouplens.org/datasets/hetrec-2011/}}
datasets.
Note that this distribution of DBLP is the bipartite graph of authors and venues, and is different from the community-based version distributed by SNAP. The LastFM dataset consists of listeners and musicians, where an edge indicates listen count, which we log-scale to improve convergence for all methods.
We start by splitting each rating set into training- and test-sets with a 40\% holdout. In the case of DBLP we use the same split as Gao et al. We use embeddings from the training bipartite graph to perform link prediction. We then compare the ranked list of training-set recommendations for each user, truncated to 10 items, to the test-set rankings.
We calculate 128-dimensional embeddings for each method, and report F1, Normalized Discounted Cumulative Gain (NDCG), Mean Average Precision (MAP) and Mean Reciprocal Rate (MRR).

\section{Significance and Impact}

In contrast to what is typically claimed in papers, we observe that the link prediction data (Table~\ref{tab:variable_holdout}) demonstrates that different graphs lead to very
different performance results for the existing state-of-the-art and proposed embeddings. Moreover, their behavior is changed with different holdouts when the size of training set is smaller. For instance, our methods are above the state of the art in the Youtube and MadGrades graphs, but 
Metapath2Vec++, Node2Vec, and LINE each have scenarios wherein they outperform 
the field. Additionally, while there are scenarios where the combination methods perform 
as expected, such as in the Youtube, MadGrades, and DBLP $B$-Personalized cases,
we observe that variability in the other proposed embeddings can disrupt 
this performance gain.
\definecolor{node2vec}{HTML}{999999}
\definecolor{metapath}{HTML}{f781bf}
\definecolor{deepwalk}{HTML}{a65628}
\definecolor{line}{HTML}{fbc02d}
\definecolor{bine}{HTML}{f57f17}
\definecolor{bhbe}{HTML}{984ea3}
\definecolor{ahbe}{HTML}{4daf4a}
\definecolor{dircomb}{HTML}{377eb8}
\definecolor{refcomb}{HTML}{e41a1c}
\def \subplotsize{0.22\linewidth}
\def \subimgsize{1\textwidth}
\begin{table}
    \centering
    \begin{tabular}{clclcl}
      {\huge\bf\color{bhbe}---}     & FOBE &
      {\huge\bf\color{ahbe}---}     & HOBE &
      {\huge\bf\color{dircomb}---}  & D.Comb. \\
      {\huge\bf\color{refcomb}---}  & A.R.Comb. &
      {\huge\bf\color{deepwalk}-~-} & Deepwalk &
      {\huge\bf\color{line}-~-}     & LINE \\
      {\huge\bf\color{node2vec}-~-} & Node2Vec &
      {\huge\bf\color{bine}-~-}     & BiNE &
      {\huge\bf\color{metapath}-~-} & Metapath2Vec++ \\
    \end{tabular}\\
    \begin{tabular}{cccc}
    \hline
    \multicolumn{1}{c}{}&
    $A$-Pers. & $B$-Pers. & Unified \\
    \rotatebox[origin=c]{90}{Amazon} &
    \begin{minipage}[c][\subplotsize]{\subplotsize}\centering
      \includegraphics[width=\subimgsize]{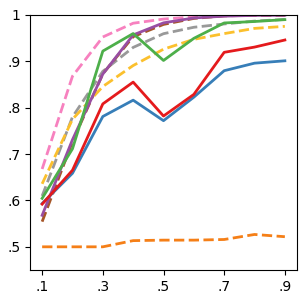}
    \end{minipage} &
    \begin{minipage}[c][\subplotsize]{\subplotsize}\centering
      \includegraphics[width=\subimgsize]{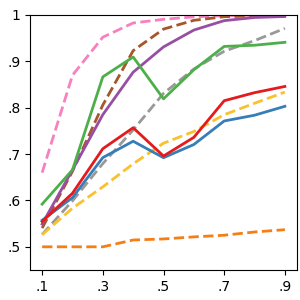}
    \end{minipage} &
    \begin{minipage}[c][\subplotsize]{\subplotsize}\centering
      \includegraphics[width=\subimgsize]{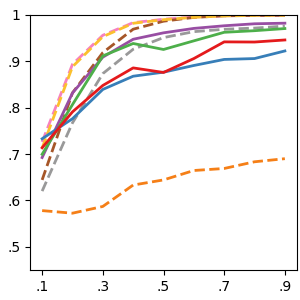}
    \end{minipage} \\
    \rotatebox[origin=c]{90}{DBLP} &
    \begin{minipage}[c][\subplotsize]{\subplotsize}\centering
      \includegraphics[width=\subimgsize]{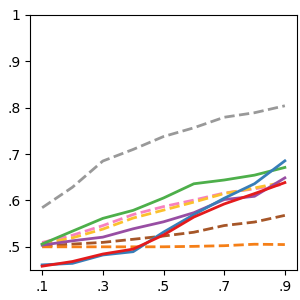}
    \end{minipage} &
    \begin{minipage}[c][\subplotsize]{\subplotsize}\centering
      \includegraphics[width=\subimgsize]{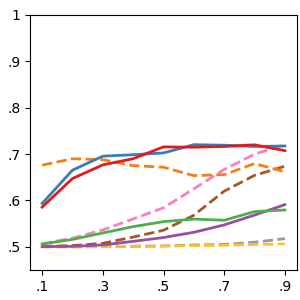}
    \end{minipage} &
    \begin{minipage}[c][\subplotsize]{\subplotsize}\centering
      \includegraphics[width=\subimgsize]{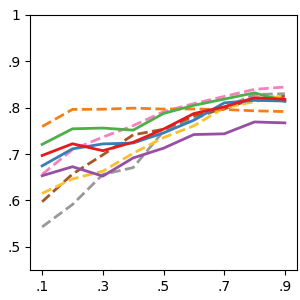}
    \end{minipage} \\
    \rotatebox[origin=c]{90}{Friendster} &
    \begin{minipage}[c][\subplotsize]{\subplotsize}\centering
      \includegraphics[width=\subimgsize]{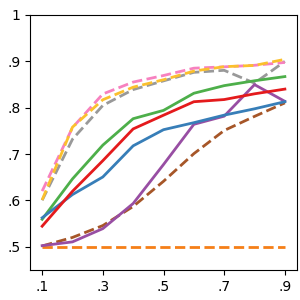}
    \end{minipage} &
    \begin{minipage}[c][\subplotsize]{\subplotsize}\centering
      \includegraphics[width=\subimgsize]{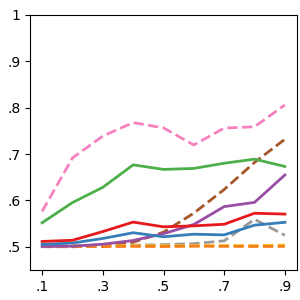}
    \end{minipage} &
    \begin{minipage}[c][\subplotsize]{\subplotsize}\centering
      \includegraphics[width=\subimgsize]{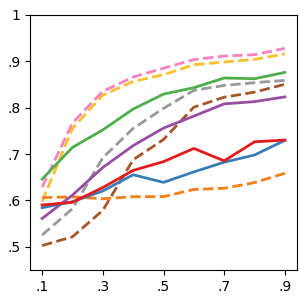}
    \end{minipage} \\
    \rotatebox[origin=c]{90}{Livejournal} &
    \begin{minipage}[c][\subplotsize]{\subplotsize}\centering
      \includegraphics[width=\subimgsize]{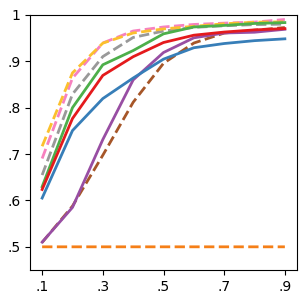}
    \end{minipage} &
    \begin{minipage}[c][\subplotsize]{\subplotsize}\centering
      \includegraphics[width=\subimgsize]{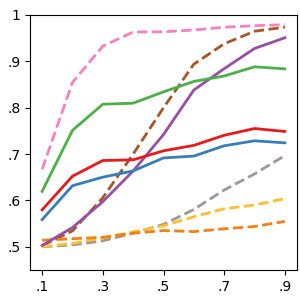}
    \end{minipage} &
    \begin{minipage}[c][\subplotsize]{\subplotsize}\centering
      \includegraphics[width=\subimgsize]{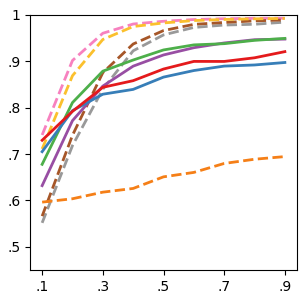}
    \end{minipage} \\
    \rotatebox[origin=c]{90}{MadGrades} &
    \begin{minipage}[c][\subplotsize]{\subplotsize}\centering
      \includegraphics[width=\subimgsize]{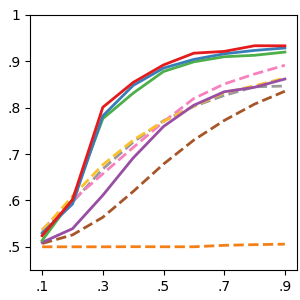}
    \end{minipage} &
    \begin{minipage}[c][\subplotsize]{\subplotsize}\centering
      \includegraphics[width=\subimgsize]{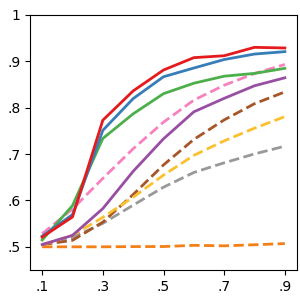}
    \end{minipage} &
    \begin{minipage}[c][\subplotsize]{\subplotsize}\centering
      \includegraphics[width=\subimgsize]{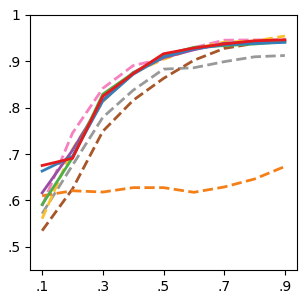}
    \end{minipage} \\
    \rotatebox[origin=c]{90}{YouTube} &
    \begin{minipage}[c][\subplotsize]{\subplotsize}\centering
      \includegraphics[width=\subimgsize]{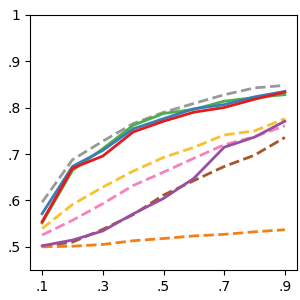}
    \end{minipage} &
    \begin{minipage}[c][\subplotsize]{\subplotsize}\centering
      \includegraphics[width=\subimgsize]{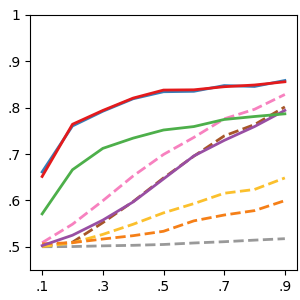}
    \end{minipage} &
    \begin{minipage}[c][\subplotsize]{\subplotsize}\centering
      \includegraphics[width=\subimgsize]{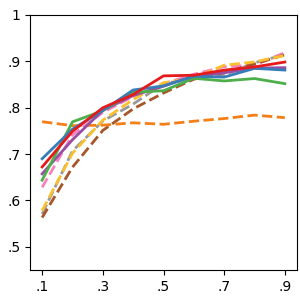}
    \end{minipage} \\
    \end{tabular}
    \caption{\label{tab:variable_holdout}
    Link Prediction Accuracy vs. Training-Test Ratio. Dashed lines indicate prior work, while solid lines indicate methods proposed here.
    }
  \end{table}

When comparing the $A$- and $B$-Personalized results, its is important to keep in mind that for all considered graphs there are more $A$ nodes ($|A| > |B|$), and therefore these nodes tend to have fewer neighbors ($\expected[\Gamma(\alpha)] < \expected[\Gamma(\beta)]$).
For this reason, we find that different embedding methods can exhibit significantly different behavior across both personalized tasks.
Intuitively, performing well on the $A$-Personalized set indicates an ability to extrapolate connections between elements with significantly more sparse attachments, such as selecting a new movie given a viewer's limited history.
In contrast, performance on the $B$-Personalized set indicates an ability to uncover trends among relatively larger sets of connections, such as determining what patterns are common across all the viewers of a particular movie.
While these two tasks are certainly related, we observe that the $B$-Personalized evaluation appears to be significantly more challenging for a number of embedding methods, such as Node2Vec on Lovejournal and YouTube. In contrast, HOBE succeeds in this evaluation for both cases, as well as Friendster and MadGrades. Metapath2Vec++ additionally is superior on LiveJournal and Friendster, but falls behind on DBLP, MadGrades, and Youtube.

In the recommendation results (Table~\ref{tab:recommendation_results_dblp}~and~\ref{tab:recommendation_results_lastfm}), our methods improve the state-of-the art. This is likely due to the behavior of aggregate neighborhood-based comparisons present within FOBE and HOBE, which has the effect of grouping clusters of nodes within one type's embedding space.
Our biggest increase is in MRR for DBLP, indicating that the first few suggestions from our embeddings are often more relevant.
The performance of HOBE, demonstrates the ability for algebraic distance to estimate useful local similarity measures. 
Interestingly, in the LastFM dataset, FOBE outperforms HOBE.
One reason for this is that LastFM contains significantly more artists-to-user than DBLP contains venues-to-author. As a result the amount of information present when estimating algebraic similarities is different across datasets, and insufficient to boost HOBE above FOBE.
\begin{table}
\small
\centering
\begin{tabular}{l|rrrr}
Metric@10:&F1&NDCG&MAP&MRR\\
\hline
DeepWalk&.0850&.2414&.1971&.3153\\
LINE&.0899&.1441&.0962&.1713\\
Node2Vec&.0854&.2389&.1944&.3111\\
MP2V++&.0865&.2514&.1906&.3197\\
BINE&{\bf.1137}&.2619&.2047&.3336\\ 
\hline
FOBE&.1108&.3771&.2382&.4491\\
HOBE&.1003&{\bf.4054}&{\bf.3156}&{\bf.6276}\\
D.Comb.&.0753&.2973&.2362&.5996\\
A.R.Comb.&.0667&.2359&.1730&.5080\\
\end{tabular}
\caption{\label{tab:recommendation_results_dblp}
DBLP Recommendation.
Note: result numbers from prior works are reproduced from~\cite{gao2018bine}.}
\end{table}

\begin{table}
\small
\centering
\begin{tabular}{l|rrrr}
Metric@10:&F1&NDCG&MAP&MRR\\
\hline
DeepWalk&.0027&.0153&.0069&.1844\\
LINE&.0067&.0435&.0229&.2477\\
Node2Vec&.0279&.1261&.0645&.2047\\
MP2V++&.0024&.0153&.0088&.2677\\
BINE&.0227&.1551&.0982&.3539\\
\hline
FOBE&{\bf.0729}&{\bf.3085}&{\bf.1997}&.3778\\
HOBE&.0195&.1352&.0789&.3400\\
D.Comb.&.0243&.1285&.0795&.3520\\
A.R.Comb.&.0388&.1927&.1249&{\bf.3915}\\
\end{tabular}
\caption{\label{tab:recommendation_results_lastfm}
LastFM Recommendations.}
\end{table}

When looking at both link prediction and recommendation tasks, we observe a highly variable performance of the combination methods. In some cases, such as the MadGrades and YouTube link prediction tasks, as well as the LastFM recommendation task, these combinations are capable of learning a joint representation from FOBE and HOBE that can improve overall performance. However, in other cases, such as the Amazon link prediction task, the combination method appears to have significantly decreased performance.
This effect is due to the increased number of hyperparameters introduced by the combination approach, which are determined not by the complexity of a given dataset, but are instead determined by the number and size of input embeddings.
In the Amazon dataset, these free parameters lead to overfitting the combination embeddings.

\section{Sensitivity Study}
We select the MadGrades network to demonstrate how our proposed methods are effected by the sampling rate. We run ten trials for each experimental sampling rate, consisting of powers of 2 from 1 to 1024. Each trial represents an independent 50\% holdout experiment. We present min, mean, and max observed link prediction accuracy.

To continue comparing FOBE and HOBE, it would appear that higher-order sampling is often 
able to produce better results, but that the algebraic distance heuristic 
introduces added variability that occasionally reduces overall performance. In 
some applications it would appear that this variability is manageable, as seen 
in our DBLP recommendation results. However in the case of link prediction on 
Amazon communities, this caused an unintentional drop when FOBE remained more consistent.
Overall, FOBE and HOBE are fast methods that broaden the array of embedding techniques available for bipartite graphs. While no method is clearly superior in every case, there exist a range of graphs and applications that are better suited by these methods.

Looking to the sensitivity study (Tables~\ref{tab:sensitivity_study}), we see the variability of HOBE is significantly larger for small sampling rates.
However, we do observe that after approximately 32 
samples per node, in the case of MadGrades, this effect is reduced.
Still, considering FOBE does not exhibit this same quality, it is likely the variability of the algebraic similarity measure that ultimately leads to otherwise unexpected reductions in HOBES performance.
\definecolor{max}{HTML}{fbc02d}
\definecolor{min}{HTML}{377eb8}
\definecolor{mean}{HTML}{e41a1c}

\def \subplotsize{0.25\linewidth}
\def \subimgsize{0.98\textwidth}
\begin{table}
    \centering
    \begin{tabular}{clclcl}
      {\huge\bf\color{max}--}  & Max &
      {\huge\bf\color{mean}--} & Mean &
      {\huge\bf\color{min}--}  & Min \\
    \end{tabular}\\
    \begin{tabular}{cccc}
    \hline
    \multicolumn{1}{c}{}&
    Per-$A$ & Per-$B$ & Unified \\
    \rotatebox[origin=c]{90}{FOBE} &
    \begin{minipage}[c][\subplotsize]{\subplotsize}\centering
      \includegraphics[width=\subimgsize]{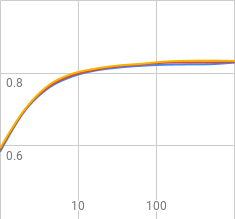}
    \end{minipage} &
    \begin{minipage}[c][\subplotsize]{\subplotsize}\centering
      \includegraphics[width=\subimgsize]{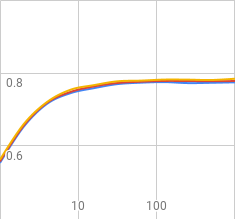}
    \end{minipage} &
    \begin{minipage}[c][\subplotsize]{\subplotsize}\centering
      \includegraphics[width=\subimgsize]{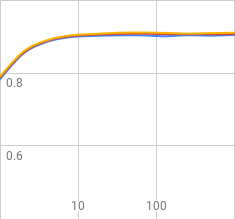}
    \end{minipage} \\
    \rotatebox[origin=c]{90}{HOBE} &
    \begin{minipage}[c][\subplotsize]{\subplotsize}\centering
      \includegraphics[width=\subimgsize]{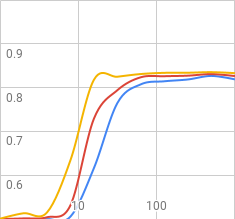}
    \end{minipage} &
    \begin{minipage}[c][\subplotsize]{\subplotsize}\centering
      \includegraphics[width=\subimgsize]{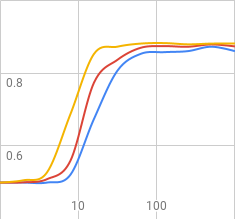}
    \end{minipage} &
    \begin{minipage}[c][\subplotsize]{\subplotsize}\centering
      \includegraphics[width=\subimgsize]{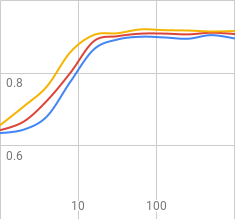}
    \end{minipage} \\
    \end{tabular}
    \caption{\label{tab:sensitivity_study}
        Link Prediction Accuracy vs. Sampling Rate. Depicts the effect of increasing $s_r$ from 2 to 1024 on the MadGrades dataset, running 10-trials of the 50\% holdout experiment per value of $s_r$.
    }
\end{table}
\section{Conclusions} \label{sec:conclusion}

In this work we present FOBE and HOBE, two strategies for modeling bipartite 
networks that are designed to capture type-specific structural properties. 
FOBE, which captures first-order relationships, samples nodes in small local 
neighborhoods. HOBE, in contrast, captures higher-order relationships that 
are prioritized by a heuristic signal provided by algebraic distance on graphs
. In addition we present two variants on an approach to learn joint 
representations that are designed to identify a ``best of both worlds'' embedding.
We evaluate these methods against the state-of-the-art via a set of link prediction and recommendation tasks.

The most novel component of FOBE and HOBE is that these methods do not encode 
cross-typed relationships through a linear transformation, but instead model 
these relationships through the aggregate behavior of node neighborhoods. For 
this reason, we find that our proposed method performs well in the context of 
recommendation, where identifying local clusters of similar nodes is important (see example of partitioning application \cite{sybrandt2019hypergraph}).
In the case of link prediction, where the goal is to identify specific 
attachments between two particular nodes, we find that the methods perform at 
a level similar to those considered in the benchmark, and only exceed the 
state-of-the-art in particular graphs.
While our personalized classification tasks demonstrate the ability for FOBE and HOBE to capture type-specific latent features, additional work is necessary to study the specific qualities these methods encode.

\begin{acks}
This work was supported by NSF awards MRI \#1725573 and NRT \#1633608.
\end{acks}

\bibliographystyle{ACM-Reference-Format}
\bibliography{main}

\appendix


\end{document}